\documentclass[review]{elsarticle}

\usepackage{amsmath,amssymb}
\usepackage{bm}
\usepackage{color}
\usepackage{ltablex}
\usepackage{booktabs}

\makeatletter
\def\@author#1{\g@addto@macro\elsauthors{\normalsize%
    \def\baselinestretch{1}%
    \upshape\authorsep#1\unskip\textsuperscript{%
      \ifx\@fnmark\@empty\else\unskip\sep\@fnmark\let\sep=,\fi
      \ifx\@corref\@empty\else\unskip\sep\@corref\let\sep=,\fi
      }%
    \def\authorsep{\unskip,\space}%
    \global\let\@fnmark\@empty
    \global\let\@corref\@empty  
    \global\let\sep\@empty}%
    \@eadauthor={#1}
}
\makeatother

\journal{Pattern Recognition}

\begin{document}

\begin{frontmatter}

\title{A Survey on Text Generation using Generative Adversarial Networks}

\author{Gustavo H. de Rosa\corref{c}}
\cortext[c]{Corresponding author}
\ead{gustavo.rosa@unesp.br}

\author{Jo\~ao P. Papa}
\ead{joao.papa@unesp.br}

\address{Department of Computing \\S\~ao Paulo State University \\ Bauru, Brazil
}

\begin{abstract}
This work presents a thorough review concerning recent studies and text generation advancements using Generative Adversarial Networks. The usage of adversarial learning for text generation is promising as it provides alternatives to generate the so-called ``natural" language. Nevertheless, adversarial text generation is not a simple task as its foremost architecture, the Generative Adversarial Networks, were designed to cope with continuous information (image) instead of discrete data (text). Thus, most works are based on three possible options, i.e., Gumbel-Softmax differentiation, Reinforcement Learning, and modified training objectives. All alternatives are reviewed in this survey as they present the most recent approaches for generating text using adversarial-based techniques. The selected works were taken from renowned databases, such as Science Direct, IEEEXplore, Springer, Association for Computing Machinery, and arXiv, whereas each selected work has been critically analyzed and assessed to present its objective, methodology, and experimental results.
\end{abstract}


\end{frontmatter}

\section{Introduction}
\label{s.introduction}

Deep Learning has received the spotlight due to its capacity to solve human-like tasks through hierarchical learning. Nevertheless, such an area depends on proper representation learning to obtain high-quality performance and accomplish the intended task. For instance, Zhang et al.~\cite{ZhangIPM:21} presented a survey on an effective paradigm for representation learning denoted as Concept Factorization, which is widely used to correlate data points and concepts as linear combinations. Furthermore, Zhang et al.~\cite{ZhangNN:21} introduced a novel approach to enhance optical character recognition with residual-based representation learning, achieving state-of-the-art results using dense Convolutional Neural Networks.

Such applications are also extensible to the Natural Language Processing (NLP) area, which also depends on precise representation learning to understand how humans communicate using voice- and text-based patterns. Essentially, its ultimate goal is to create algorithms capable of interpreting the human language and between machines and humans.

A common task NLP-based task denoted as Language Modeling aims to understand the general text structure by learning its grammatical, syntactic, and semantical rules. From a computational point of view, this task aims at predicting a $t+1$ timestep based on previous $t$ timesteps, where each timestep can be composed of characters, words, or even sentences. Thus, a well-learned language model can reproduce the training text structure by predicting new data based on past information, e.g., text generation. For example, imagine that a fictitious language composed of three tokens follows this structure: \textless subject\textgreater~\textless verb\textgreater~\textless noun\textgreater. If a language model is capable of learning such structure, it is also capable of predicting a \textless verb\textgreater~token based on input \textless subject\textgreater~token, as well as predicting a \textless noun\textgreater~token based on input \textless subject\textgreater~and \textless verb\textgreater~tokens.

One may realize that such a task is notably different from traditional Machine Learning ones as it employs a new dependency, i.e., the time itself. In other words, tokens (characters, words, sentences) are interpreted as sequences, where each piece represents a particular timestep. Additionally, previous sequences must be carried to the future during learning, as the language model predicts data based on past information~\cite{Mikolov:10}. Traditionally, researchers focus on solving such a problem using Recurrent Neural Networks (RNN)~\cite{Elman:90}, which can re-use prior knowledge to learn future information. Even though RNNs seem to be the ideal structure to tackle text generation problems, they are still susceptible to adversarial manipulation~\cite{Biggio:18}. In other words, such models are vulnerable to slight modifications in data that were not introduced in the learning process, resulting in mispredictions and inefficient performance. This vulnerability represents potential security threats and may compromise real-world applications, remaining as an open research field.

Goodfellow et al.~\cite{Goodfellow:14} proposed the Generative Adversarial Networks (GAN) and achieved several hallmarks concerning adversarial learning. Some recent works~\cite{Hu:19, Li:20, LLiu:20} revealed their capability of learning data distributions and plausibly generating artificial image-based data. Notwithstanding, such networks were designed to deal with continuous information instead of discrete data~\cite{Hjelm:18}, i.e., they are not suitable when applied with sequences of texts. Thus, several works attempt to tackle such an issue by employing Gumbel-Softmax differentiation~\cite{Jang:16}, Reinforcement Learning (RL)~\cite{Sutton:98}, or modified training objectives~\cite{Che:17}. This paper proposes to survey and discuss the approaches mentioned above applied explicitly to text generation tasks. Additionally, it aims to provide a comprehensive source of text-based GAN advancements, where architectures are critically analyzed in terms of datasets, objectives, evaluation metrics, and experimental results.

The remainder of this paper is organized as follows. Section~\ref{s.survey} presents the survey structure, related surveys, and how surveyed works were selected. Section~\ref{s.background} presents some theoretical background concerning Natural Language Processing, Language Modeling, Generative Adversarial Networks, as well as some text generation datasets and topmost evaluation metrics. Section~\ref{s.analysis} discusses and provides an in-depth analysis of the surveyed works. Finally, Section~\ref{s.conclusion} states overall conclusions and future insights regarding the surveyed context.
\section{Survey Structure}
\label{s.survey}

\subsection{Related Surveys}
\label{ss.related_surveys}

Text generation, formerly known as natural language generation, has received a great deal of attention throughout the last years, mainly due to the advance of computational power and Deep Learning-based architectures. Gatt et al.~\cite{Gatt:18} presented a survey regarding the advancements of natural language generation. They categorized and discussed state-of-the-art data-driven architectures and presented the main drawbacks related to the synergy between natural language generation and other Artificial Intelligence areas. Nevertheless, their survey aims to provide an overview of the task instead of diving into the models' architecture and their advantages/disadvantages.

A recent survey presented by Lu et al.~\cite{Lu:18} reviews and critically analyzes adversarial-based text models' development. Even though their survey only focuses on comparing six state-of-the-art architectures and the Maximum Likelihood Estimation over two benchmarking datasets, the authors established a reference for those wanting to work with text-based adversarial learning. Furthermore, Zhang et al.~\cite{Zhang:20} proposed a survey in the context of adversarial attacks on Deep Learning models in Natural Language Processing, where they targeted on summarizing and discussing the types of adversarial attacks instead of discussing the latest advancements regarding Generative Adversarial Networks applied to text-based generation.

Therefore, the presented survey aims to fulfill the literature's blank spots and provide a unique reference of adversarial-based models applied to text generation. Extensive research has been conducted across several renowned databases to identify the most promising text-based Generative Adversarial Networks. Additionally, the selected models have been critically analyzed and discussed in terms of objective, methodology, experimental results, and evaluation metrics. Finally, the present work intends to categorize the text-based adversarial models according to their differentiability strategy\footnote{It is essential to remember that Generative Adversarial Networks were not designed to work with discrete data.}, e.g., Gumbel-Softmax differentiation, Reinforcement Learning, and modified training objectives.

\subsection{Paper Selection}
\label{ss.paper_selection}

Initially, we performed an extensive search at renowned databases, such as Science Direct, IEEEXplore, Springer, Association for Computing Machinery (ACM), and arXiv, to define the starting year of the surveyed works. As the Generative Adversarial Networks had been proposed in 2014 and the Gumbel-Softmax approach had been proposed in 2016, we could only find GAN-based text generation with Gumbel-Softmax references beyond 2016. Additionally, no work addressed GAN-based text generation with Reinforcement Learning and modified training objectives before 2017.

\begin{figure}[!htb]
    \centering
    \includegraphics[scale=0.5]{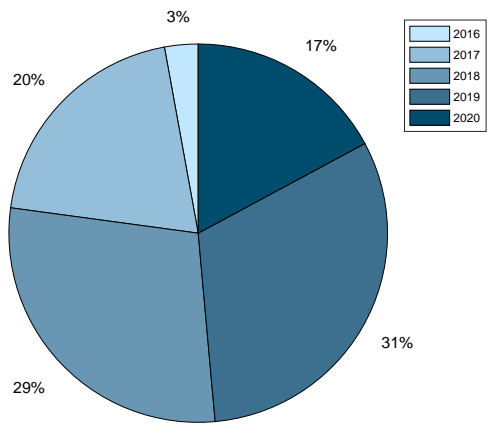}
    \caption{Percentage of considered publications per publishing period.}
    \label{f.pub_year}
\end{figure}

With that in mind, we considered two keywords for performing the search: (i) ``generative adversarial network", and (ii) ``text generation", and gathered the resulting works that have been cited at least one time and that were relevant to this survey context (GAN-based text generation). Furthermore, we verified their authors' background to check whether they were active researchers from truthful laboratories and universities. Thus, the following publication period is covered by the current survey: 2016 (1 work), 2017 (7 works), 2018 (10 works), 2019 (11 works), and 2020 (6 works), as illustrated by Figure~\ref{f.pub_year}.
\section{Theoretical Background}
\label{s.background}

This section presents a brief theoretical background regarding Natural Language Processing, Language Modeling, Generative Adversarial Networks, Reinforcement Learning, and Gumbel-Softmax differentiation-based strategies. Additionally, it describes an overview concerning the four most commonly employed text generation datasets and some topmost evaluation metrics.

\subsection{Natural Language Processing}
\label{ss.nlp}

Most human communication is performed through voice and text, i.e., vital sources of information used to predict behavioral characteristics. Nevertheless, it is hard to predict human-based communication because each individual possesses particular traits in its written and verbal language, e.g., pitch, word selection, vocabulary, and accent. 

An area that arises in tackling such a problem is Natural Language Processing, responsible for modeling and comprehending the interaction between humans and machines through natural language usage. Deep Learning has recently achieved several hallmarks regarding NLP-based tasks~\cite{Goldberg:16}, introducing state-of-the-art models for a wide variety of tasks, such as language modeling, grammatical analysis, and named entity recognition, among others.

Nonetheless, one may perceive that NLP is often complicated due to language's intrinsic features (morphology, syntax, semantic) and the requirement of knowledge to connect sequences of characters and words. Even though humans quickly learn and master languages, machines still struggle in such a task, leaving it as an open research problem for the community.
 
\subsubsection{Language Modeling}
\label{sss.language}

Language Modeling tasks stand for learning probabilistic models that can predict a $t+1$ timestep given $t$ timesteps, e.g., predict the next word given a sequence of words. A language model learns the probability of token occurrences based on examples, where simpler models might look at sequences of characters or words, while larger models may learn sequences of sentences or paragraphs. Notwithstanding, such tasks are often complicated as models need to learn natural languages' underlying traits, such as grammar, syntax, and semantic. Additionally, as its learning procedure is based on a set of examples, it poorly performs when employed with unknown data, e.g., not-learned tokens.

Most language models are used as root architectures for larger NLP tasks, such as speech recognition, optical character recognition, machine translation, and spelling correction. On the other hand, a trained language model can generate new text according to the structure it has already learned. Recently, most language models are fostered by neural networks, achieving state-of-the-art results due to their capacity for generalization. Furthermore, it is common to use pre-trained embeddings, i.e., real-valued vectors that encode high-dimensional representations of tokens (characters or words) in a projected vector space, to tackle the curse of dimensionality\footnote{Curse of dimensionality happens when vast vocabularies are used, producing sparse representations.}.

One may perceive that an interesting approach is to use RNNs, as they can re-use their previous hidden states, where neurons are fed with recurrent connections and learn a so-called ``memory". For example, a feed-forward network is not fit to create a probabilistic distribution capable of knowing that a \textless noun\textgreater~should come after a \textless verb\textgreater, as it only looks into the ``future". On the other hand, recurrent networks can look into their past ``memory"~and decide their subsequent output based on previous inputs.

\subsubsection{How to Generate Text?}
\label{sss.text}

\begin{sloppypar}
A properly trained language model encodes features and rules from the text it was trained. In other words, the model has learned a probabilistic distribution that represents the training data and samples new data. Let $P(w_t | w_{t-1}, w_{t-2}, \ldots, w_{t-n})$ be the probability distribution of a token at $t$ timestep, given an $n$-sized sequence of previous tokens. The language model estimates the probabilities of a token $w_t$ given a previous sequence of tokens, allowing it to be sampled into a new token. Afterward, in $t+1$ timestep, the probability distribution tries to represent $w_{t+1}$ given $w_{t}, w_{t-1}, \ldots, w_{t-n+1}$. The process is repeated until a convergence criterion is satisfied, such as the number of generated tokens. Finally, the resulting tokens are sequentially arranged to form the generated text.
\end{sloppypar}

\subsection{Generative Adversarial Networks}
\label{ss.gan}

\begin{figure}[!htb]
    \centering
    \includegraphics[scale=0.4]{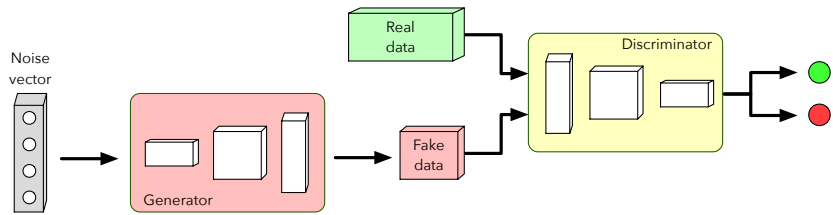}
    \caption{Standard architecture of a Generative Adversarial Network.}
    \label{f.gan}
\end{figure}

Szegedy et al.~\cite{Szegedy:13} found in their work that several Machine Learning architectures, including state-of-the-art, are vulnerable to adversarial manipulation, i.e., these models could not correctly classify when fed with slightly different learning data. Years ago, Goodfellow et al.~\cite{Goodfellow:14} introduced the Generative Adversarial Networks as an alternative to the adversarial problem. It is implemented as a two-neural networks system, where the networks compete in a zero-sum approach. Namely, the idea is to have a discriminative and a generative network, where the generative one is in charge of producing fake data, and the discriminative one is in charge of estimating the probability of the fake data being real. Figure~\ref{f.gan} illustrates an example of a standard Generative Adversarial Network.

The network's discriminator $D$ is trained to maximize the probability of classifying both training and generated data as real images. Simultaneously, the network's generator $G$ is trained to minimize $log(1 - D(G(\bm{z}))$, i.e., the divergence among both data distributions, where $\bm{z}$ denotes a noisy input. In other words, the two neural-networks system compete in a zero-sum game, represented by Equation~\ref{e.gan_min_max}:

\begin{equation}
\label{e.gan_min_max}
\min_G \max_D C(D,G) = \mathbb{E}_{\bm{x}}[logD(\bm{x})] + \mathbb{E}_{\bm{z}}[log(1-D(G(\bm{z}))],
\end{equation}
where $C(D,G)$ is the loss function to be minimized, $D(\bm{x})$ is the discriminator's estimated probability of a real sample $\bm{x}$ being real, $\mathbb{E}_{\bm{x}}$ is the mathematical expectancy over all samples from the real data set $\mathcal X$, $G(\bm{z})$ stands for the generated fake data given the noise vector $\bm{z}$, $D(G(\bm{z}))$ is the estimated probability of a fake sample $G(\bm{z})$ being real, and $\mathbb{E}_{\bm{z}}$ is the mathematical expectancy over all random generator inputs, i.e., the expected value over all fake samples generated by $G$.

Nevertheless, a deriving problem from Equation~\ref{e.gan_min_max} is that GANs can get trapped in local optimums when the discriminator has an easy task. In other words, at the beginning of the training, when $G$ still does not generate proper samples, $D$ can reject fake samples with high probability, as they are incredibly different from the real data\footnote{Such a problem saturates the function $log(1-D(G(\bm{z}))$.}. Thus, an alternative to such a problem is to train $G$ to maximize $log(D(G(\bm{z}))$, which improves the initial gradient values.
    
\subsubsection{Differentiating GANs with Gumbel-Softmax}
\label{sss.gumbel}

During GANs training, both generator and discriminator need a differentiable loss function to update their parameters. However, when using RNN-based architectures as GANs generators, it is impossible to differentiate its output, usually accomplished by sampling one-hot categorical values from a Softmax distribution.

A solution to such a problem proposed by Jang et al.~\cite{Jang:16} consists of sampling from the so-called Gumbel-Softmax, which is a continuous distribution that can approximate samples from categorical distributions. Essentially, the idea is to employ the Softmax distribution along with a Gumbel distribution and a temperature parameter, as follows:

\begin{equation}
\label{e.gumbel_softmax}
\bm{\tilde{y}} = \frac{e^{\frac{\bm{w}+\bm{g}}{\tau}}}{\sum e^{\frac{\bm{w}+\bm{g}}{\tau}}},
\end{equation}
where $\bm{\tilde{y}}$ stands for the differentiable outputs, $\bm{w}$ is a vector of unnormalized log probabilities, $\bm{g}$ is a vector sampled out of the Gumbel distribution, and $\tau$ is the temperature parameter. As $\tau \rightarrow 0$, $\bm{\tilde{y}}$ approximates a one-hot categorical distribution, while $\bm{\tilde{y}}$ approximates a uniform distribution as $\tau \rightarrow \infty$.

\subsubsection{Reinforcing GANs Learning}
\label{sss.rl}

As previously mentioned, GANs need a differentiable loss function in order to perform their training. An alternative to the Gumbel-Softmax trick consists of creating a Reinforcement Learning-based agent for the generator. Essentially, an RL algorithm is defined by a policy function $\pi(\bm{a} | \bm{s}, \theta)$ parameterized by $\theta$, which outputs a probability distribution over all actions $\bm{a}$ given the current state $\bm{s}$ of an agent. Note that $\bm{s}$ stands for the state of the agent (text generated), and $\bm{a}$ stands for its actions (text to be generated). In other words, an agent generates text based on the policy $\pi$, where its actions are sampled according to the distribution defined by the same policy. Therefore, it is possible to optimize $\theta$ by employing policy gradient methods, such as the REINFORCE algorithm~\cite{Williams:92}, to build a good text generation agent.

Let $\bm{y} = (y_1, y_2, \ldots, y_T) \mid \bm{y} \in \mathcal{Y}$ depict the sequence generation problem, where $\mathcal{Y}$ is the vocabulary of candidate tokens and $T$ is the number of tokens to be generated. Let the generative model $G$ be parametrized by $\theta_G$ and generate a state $s$ for each time $t$, which corresponds to a generated token $y_t$. Further, let its next action, denoted by $a$, be the next token $y_{t+1}$ to be chosen. Therefore, the policy $G(y_t\mid\bm{y}_{1:t-1}; \theta_G)$ is stochastic, such that the transition state is deterministic after an action has been chosen, i.e., $\delta_{s,\tilde{s}}^{a} = 1$ for the next state $\tilde{s} = \bm{y}_{1:t}$ if the current state $s = \bm{y}_{1:t-1}$ and the action $a = y_t$.

Additionally, the discriminative model $D$, which is parameterized by $\theta_D$, guides the generator through a probability $D(\bm{y})$ that indicates whether a sequence $\bm{y}$ belongs to real data or not. This discriminative model is trained by positive (real data) and negative (artificial data) samples, providing a reward signal to update the generative model's gradients. Equation~\ref{e.seqgan_policy} describes the reward signal obtention procedure:

\begin{equation}
\label{e.seqgan_policy}
Q^G_D(a;s) =
	\left.
	\begin{cases}
		\frac{1}{N} \sum_{n=1}^N D(\bm{y}^n) \mid \bm{y}^n \in MC(\bm{y}_{1:t}; N) & \text{for } t < T\\
		D(\bm{y}) & \text{for } t = T
	\end{cases}
	\right.,
\end{equation}
where $Q^G_D(a;s)$ is the action-value function of a sequence, i.e., the reward signal accumulated from state $s$ given an action $a$, and $N$ is the number of Monte Carlo samplings.

\subsection{Datasets}
\label{ss.datasets}



The Amazon Customer Reviews\footnote{\url{https://nijianmo.github.io/amazon/index.html}}~\cite{Ni:19} dataset, commonly denoted as Product Reviews, is composed of more than $233,000,00$ users reviews, which have expressed their opinion and described their experience amongst products bought over Amazon's website. Furthermore, the Chinese Poems\footnote{\url{https://homepages.inf.ed.ac.uk/mlap/Data/EMNLP14}}~\cite{Zhang:14} dataset is composed of $284,299$ poems from several online sources, such as Tang Poems, Song Poems, Song Ci, Ming Poems, Qing Poems, and Tai Poems.

The COCO Image Captions\footnote{\url{https://cocodataset.org}}~\cite{Chen:15} dataset has been developed through the capture of image captions belonging to the Microsoft Common Objects in Context (MS COCO) dataset. In sum, humans created $1,026,459$ captions from the same data sets used by MS COCO and divided them across two sub-sets: one with $5$ reference sentences for each image and another one with $40$ reference sentences for $5,000$ randomly chosen images. All candidate captions have been pre-processed and tokenized according to the Stanford PTBTokenizer, following the same protocols established by the Penn Treebank dataset. After pre-processing, all punctuation marks have been removed to standardize the captions. Finally, the EMNLP2017 WMT News\footnote{\url{http://statmt.org/wmt17/results.html}}~\cite{Ondrej:17} dataset has been developed to foster machine translation tasks, being composed of pairs of documents, such as sets of sentences amongst two distinct languages. Nevertheless, it is possible to adjust such a dataset to the natural language generation task and use only sentences corresponding to the same language.

\subsection{Evaluation Metrics}
\label{ss.metrics}

The task of natural language generation allows the machine to create artificial information and understand natural languages. However, it is necessary to assess such information's quality and compare them against those created by human beings. Therefore, the advent of algorithms focused on evaluation metrics allowed the estimation of how ``relevant"~the generated information is and enabled ways to benchmark their quality between several language generation architectures.

\subsubsection{Bilingual Evaluation Understudy}
\label{sss.bleu}

Bilingual Evaluation Understudy (BLEU)~\cite{Papineni:02} evaluates the quality of machine-generated text and is structured by the correspondence of source text (written by humans) and artificial text (generated by machines). Essentially, it is calculated by the co-occurrence of $n$-grams between the reference and candidate sentences, computing a type of precision between the sentences. Let $\Phi = \{{\phi}_1, {\phi}_2, \ldots, {\phi}_{\kappa}\}$ be the reference sentences with $\kappa$ tokens, while $\tilde{\Phi} = \{\tilde{\phi}_1, \tilde{\phi}_2, \ldots, \tilde{\phi}_{\omega}\}$ is the candidate sequence with $\omega$ tokens. Equation~\ref{e.bleu} describes its formula, as follows:

\begin{equation}
\label{e.bleu}	
P_n = \frac{\sum_{k} \min (d_k(\tilde{\phi}), \max d_k(\phi))}{\sum_{k} d_k(\tilde{\phi})},
\end{equation}
where $d_k(\cdot)$ is the amount of occurrences of an $n$-gram and $k$ is the maximum number of possible $n$-grams. However, such formula inflates the metric's value for short sentences and needs a penalization factor according to Equation~\ref{e.bleu_penalty}:

\begin{equation}
\label{e.bleu_penalty}	
b = \left.
\begin{cases}
1 & \text{if } \omega > \kappa\\
e^{1 - \frac{\kappa}{\omega}} & \text{if } \omega \leq \kappa
\end{cases}
\right..
\end{equation}

Therefore, the final BLEU score is calculated by a geometric mean weighted by the individual's $n$-grams precision. Moreover, the metric's value is within the $[0, 1]$ interval, where values close to $1$ indicate a bigger similarity between the sentences. Equation~\ref{e.bleu_score} describes such procedure, as follows:

\begin{equation}
\label{e.bleu_score}	
B_N = b \cdot \exp((\sum_{n=1}^{N}w_n \log(P_n))),
\end{equation}
where $n = 1, 2, 3, \ldots, N$ and $w_n$ is a constant for all $n$ values.

\subsubsection{Negative Log-Likelihood}
\label{sss.nll}

Negative Log-Likelihood (NLL)~\cite{Myung:03} is usually used as a loss function in Machine Learning training algorithms and aims at maximizing the probability of correct classifications. In other words, during a linguistic model training, NLL is used to verify whether the network is learning to generate $t+1$ tokens given the previous $t$ tokens.

Let $K$ be the number of classes to be classified, i.e., the number of possible generated tokens. The loss function is defined according to Equation~\ref{e.nll}:

\begin{equation}
\label{e.nll}
NLL = -\frac{1}{n} \; \sum_{\bm{x} \in \mathcal{X}} \; \sum_{k=1}^{K} y_k \cdot e^{a_k},	
\end{equation}
where $n$ is the number of samples in training set $\mathcal{X}$, $\bm{x}$ is the input data, $\bm{y}$ is the true label in one-hot encoding, and $a$ is the network's output.  The NLL metric is commonly used to evaluate a network's learning quality and is restricted to a positive interval, e.g., $[0, +\infty]$. Additionally, as NLL tends to $0$, the network can learn the training set patterns, i.e., correctly predicting the training samples' labels.

\subsubsection{Perplexity}
\label{sss.perplexity}

Perplexity (PPL)~\cite{Brown:92} is calculated by the exponentiation of NLL, furnishing a more intuitive metric. Given that PPL is just the exponentiation of NLL, its interval is restricted to $[1, +\infty]$, where smaller values correspond to a well-trained network, and higher values correspond to a poor learning procedure.



%
%
\section{Analysis and Discussion}
\label{s.analysis}

\begin{figure}[!htb]
    \centering
    \includegraphics[scale=0.3]{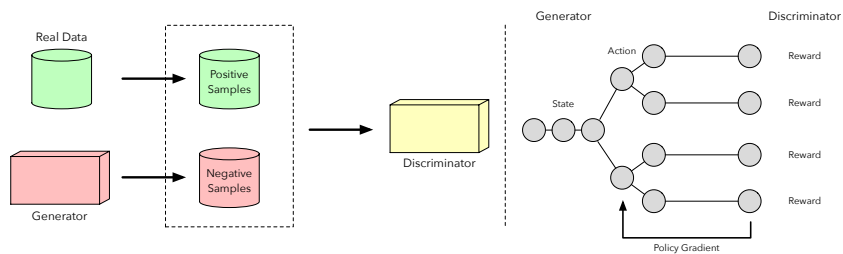}
    \caption{Pipeline commonly used to conduct the adversarial-based text generation. Note that for Gumbel-Softmax differentiation and modified training objectives, the discriminator directly rewards the generator (without Reinforcement Learning) as their functions are differentiable.}
    \label{f.pipeline}
\end{figure}

This section presents the critical analysis and discussion regarding the compiled works, which are categorized as follows: (i) GAN-based text generation with Gumbel-Softmax differentiation, (ii) GAN-based text generation with Reinforcement Learning, and (iii) GAN-based text generation with modified training objectives. Additionally, Figure~\ref{f.pipeline} describes the methodological pipeline used to conduct text generation with the mentioned approaches, and Table~\ref{t.summary} describes an overview regarding the surveyed works, where the following columns have depicted each work: reference, architecture, datasets, objectives, evaluation metrics, and experimental results, while Table~\ref{t.benchmark_bleu} comprehends a set of architectures evaluated with the most common metric (BLEU-2) over the four most common datasets: Amazon Reviews, Chinese Poems, COCO Image Captions, and EMNLP2017 WMT News.

\begin{center}	
	\footnotesize
	\setlength\LTleft{-4.25cm}
	\begin{tabularx}{1.7\linewidth}
	 {XXXXXX} 
        \\ \toprule \endfirsthead
		\toprule \endhead
		\midrule\multicolumn{6}{r}{\itshape continues on next page}\\\midrule\endfoot
	    \bottomrule
	    \caption{Summarization of the works considered in this survey.}
	   	\label{t.summary}
	    \endlastfoot
		\textbf{Reference} & \textbf{Architecture} & \textbf{Dataset} & \textbf{Objectives} & \textbf{Metrics} & \textbf{Results}\\ \midrule
		Kusner, Hernández-Lobato~\cite{Kusner:16} & GSGAN & Context-Free Grammar & Gumbel-Softmax with discrete data. & Visual inspection & Samples similar to an MLE-trained LSTM. \\ \midrule
		Nie et al.~\cite{Nie:19} & RelGAN & COCO Image Captions, EMNLP2017 WMT News & Relational Memory generator and Gumbel-Softmax distribution. & NLL, BLEU & Outperformed models when RelGAN pre-training was good. \\ \midrule
		Yin et al.~\cite{Yin:20} & Meta-CoTGAN & COCO Image Captions, EMNLP2017 WMT News & Cooperative approach with an additional language model. & NLL, BLEU & Positive impact of cooperatively approach and meta optimization. \\ \midrule
		Yu et al.~\cite{Yu:17} & SeqGAN & Chinese Poems, Nottingham Music, Obama Speech, Synthetic Data & Stochastic Reinforcement Learning gradient policy updates. & NLL, BLEU & First model to extend GANs to discrete token generation.\\ \midrule
		Lin et al.~\cite{Lin:17} & RankGAN & Chinese Poems, COCO Image Captions, Shakespeare Plays & Ranked data instead of real and fake. & BLEU & Could not compete with human-written sentences.\\ \midrule
		Wang et al.~\cite{Wang:17} & VGAN & Taobao Reviews, Amazon Review, Penn Treebank & High-latent variables with Variational AutoEncoders (VAE). & NLL, BLEU & Successfully applied VAEs to adversarial-based models. \\ \midrule
		Guimaraes et al.~\cite{Guimaraes:18} & ORGAN & Molecules Dataset, Essen Associative Code & Domain-specific objectives and discriminator's rewards. & Diversity, Validity & Generated domain-based data, with RL being essential.\\ \midrule
		Guo et al.~\cite{Guo:18} & LeakGAN & Chinese Poems, COCO Image Captions, EMNLP2017 WMT News, Synthetic Data & Leaked high-level features from the discriminator into the generator. & NLL, BLEU & Concluded that leaked information is vital to the network's learning.\\ \midrule
		Hjelm et al.~\cite{Hjelm:18} & BGAN & Google One Billion Words & Compute importance weights with difference measures. & Visual Inspection & Scratch stable training yet poor generation.\\ \midrule
		Fedus et al.~\cite{Fedus:18} & MaskGAN & IMDB Movie, Penn Treebank & Actor-critic conditional architecture. & Percentage of Unique $n$-Grams & Produced reasonable sentences, yet not as good as the ground truth.\\ \midrule
		Li et al.~\cite{Li:18} & CS-GAN & Amazon Review, Emotion, NEWS, Stanford Sentiment Treebank, Yelp Review & Incorporated sentence category over the architecture. & NLL, Accuracy & Suffered from large amounts of data and binary labels.\\ \midrule
		Balagopalan et al.~\cite{Balagopalan:18} & RE-GAN & Context-Free Grammar & Comparative study between three distinct policy gradient methods. & Goodness Score & Needed a significant pre-training.\\ \midrule
		Shi et al.~\cite{Shi:18} & IRL & COCO Image Captions, IMDB Movie, Synthetic Data & Inverse RL architecture. & BLEU, F-BLEU, B-BLEU, HM-BLEU & Alleviated the problem of reward sparsity and mode collapse.\\ \midrule
		Xu et al.~\cite{Xu:18} & DP-GAN & Amazon Review, OpenSubtitles Dialogue, Yelp Review & Novel discriminator to distinguish between repeated and novel text. & BLEU, Relevance, Diversity, Fluency & Produced more diverse text, and better distinguished novel and repeated text.\\ \midrule
		Chen et al.~\cite{Chen:19} & CTGAN & Amazon Review, Film Review, Obama Speech, Yelp Review & Mimicked human-like text generation by introducing new features. & Similarity, Diversity, Word Frequency, Subjectivity, Sentiment & Simplified text generation process outperformed baselines.\\ \midrule
		Wang, Wan~\cite{Wang:19} & SentiGAN, C-SentiGAN & BeerAdvocate, Customer Reviews, Emotional Tweet, Stanford Sentiment Treebank & Multiple generators with a penalty-based objective. & Novelty, Diversity & Tagged as promising approaches in domain-based text generation.\\ \midrule
		Zhang et al.~\cite{Zhang:19} & CD-GAN & Amazon Review, Yelp Review & Annotated data from distinct domains to generated emotional text. & Sentiment Transfer Strength, Domain Transfer Strength, Cosine Similarity, Word Overlap & Achieved comparable results to the compared baselines (sentiment transfer state-of-the-art).\\ \midrule
		Zhang et al.~\cite{CZhang:19} & TranGAN & Penn Treebank & Transformer-based architecture trained with an actor-critic RL algorithm. & Perplexity, Percentage of Unique $n$-Grams, BLEU & Processed discrete sequences and outperformed compared baselines.\\ \midrule
		Liu et al.~\cite{Liu:19} & BFGAN & Amazon Review, Chinese Sentence Making Corpus, DailyDialog & Backward and forward generators with a discriminator capable of guiding their joint training. & BLEU, SLEU & Improved quality of lexically-constrained sequences, and did not need additional labels.\\ \midrule
		Sun et al.~\cite{Sun:19} & QuGAN & Tibetan Q\&A & QRNNs and improved both reward and Monte Carlo search strategies. & BLEU & Model with optimized Monte Carlo search and BERT outperformed compared models.\\ \midrule
		Rizzo, Van~\cite{Rizzo:20} & C-SeqGAN & Amazon Review, COCO Image Captions, EMNLP2017 WMT News & Context adaption of global word embeddings, as well as a self-attention discriminator. & BLEU, Pos-BLEU, Self-BLEU & Improved SeqGAN by using global knowledge adapted to the dataset's domain.\\ \midrule
		Wu, Wang~\cite{Wu:20} & TG-SeqGAN & COCO Image Captions & Rewarded the generator with the distance between real and fake data. & NLL, Embedding Similarity & Improved the training convergence and the text quality.\\ \midrule
		Zhou et al.~\cite{Zhou:20} & SAL & COCO Image Captions, EMNLP2017 WMT News, Synthetic Data & Comparative discriminator instead of a binary one, as well as a self-improvement reward mechanism. & NLL, BLEU, Perplexity, Frechet Distance & Improved the rewards system by reducing the sparsity and mode collapse, leading to a better text generation.\\ \midrule
		Che et al.~\cite{Che:17} & MaliGAN & Chinese Poems & Normalized maximum likelihood with importance sampling and variation reduction. & BLEU, Perplexity & Created distributions with lower variances and a better discriminator.\\ \midrule
		Zhang et al.~\cite{Zhang:17} & TextGAN & Combined data from BookCorpus and ArXiv & High-dimensional latent feature distribution with kernelized discrepancy. & BLEU, Number of Kernel Density Estimators & Learned a suitable latent representation space and produced realistic sentences.\\ \midrule
		Gulrajani et al.~\cite{Gulrajani:17} & WGAN-GP & Google One Billion Words & Gradient penalization. & Visual Inspection & Generated discrete data, yet not comparable to baselines.\\ \midrule
		Press et al.~\cite{Press:17} & AltGAN & Google One Billion Words & Curriculum learning and slow teaching. & Percentage of Unique $n$-grams & Adequately trained from scratch.\\ \midrule
		Chen et al.~\cite{Chen:18} & FM-GAN & COCO Image Captions, CUB Captions, EMNLP2017 WMT News & Latent feature distributions with Feature-Mover's Distance. & Test-BLEU, Self-BLEU, Human Evaluation & Produced good sentences, yet still suffered from mode collapse.\\ \midrule
		Donahue, Rumshisky~\cite{Donahue:18} & LaTextGAN & Toronto Book Corpus & AutoEncoder. & Human Discriminator, BLEU, t-SNE Plots & t-SNE plots showed that it could model the latent space.\\ \midrule
		Li et al.~\cite{Li:19} & JSD-GAN & Chinese Poems, COCO Image Captions, EMNLP2017 WMT News, Obama Speech & Optimize Jensen-Shannon Divergence. & BLEU & RL-free with direct generator optimization, yet only for explicit representations.\\ \midrule
		Haidar et al.~\cite{Haidar:19} & TextKD-GAN & Google One Billion Words, Stanford Natural Language Inference & Knowledge distillation technique based on teacher-student model. & BLEU & Smoother text representations and outperformed traditional GANs. \\ \midrule
		Ahamad~\cite{Ahamad:19} & STGAN & BookCorpus, CMU-SE & Encoder-decoder architecture inspired by the skip-gram model. & BLEU & Skip-thought embeddings could generate discrete data.\\ \midrule
		Montahaei et al.~\cite{Montahaei:19} & DGSAN & Chinese Poems, COCO Image Captions, EMNLP2017 WMT News & Optimized closed-form solution to find an optimal discriminator. & NLL, BLEU, Self-BLEU, MS-Jaccard & Outperformed baselines regarding the MS-Jaccard metric, yet achieved slightly lower Self-BLEU.\\ \midrule
		Yang et al.~\cite{Yang:20} & FGGAN & Chinese Poems, COCO Image Captions, Synthetic Data & Feature guidance module and a new vocabulary mask to better represent semantic rules. & NLL, BLEU & Outperformed current baselines and generated more realistic sentences due to token restriction.\\ \midrule
		Wu et al.~\cite{QWu:20} & TextGAIL & COCO Image Captions, EMNLP2017 WMT News & Pre-trained language models with contrastive discriminator and proximal policy optimization. & NLL, BLEU, Self-BLEU, Perplexity & Generated diverse and accurate outputs due to the incorporation of pre-trained GPT-2 and pre-trained RoBERTa.
	\end{tabularx}
\end{center}

\begin{table}[!h]
	\footnotesize
	\renewcommand{\arraystretch}{1}
	\setlength\LTleft{-0.8cm}
	\centering
	\begin{tabularx}{\textwidth}{ccccc}
		\\ \toprule
		\textbf{Architecture} & \textbf{Amazon Review} & \textbf{Chinese Poems} & \textbf{COCO Captions} & \textbf{WMT News}
		\\ \midrule
		MLE & $-$ & $0.667$ & $0.781$ & $0.768$
		\\
		RNNLM & $0.848$ & $-$ & $-$ & $-$	
		\\ \midrule
		RelGAN & $-$ & $-$ & $0.849$ & $0.881$
		\\
		Meta-CoTGAN & $-$ & $-$ & $0.858$ & $0.882$
		\\ \midrule
		SeqGAN & $0.856$ & $0.739$ & $0.745$ & $0.777$
		\\
		VGAN & $0.868$ & $-$ & $-$ & $-$
		\\
		RankGAN & $-$ & $0.812$ & $0.743$ & $0.727$
		\\
		LeakGAN & $-$ & $\mathbf{0.881}$ & $0.746$ & $0.826$
		\\
		IRL & $-$ & $-$ & $0.829$ & $-$
		\\
		BFGAN & $\mathbf{0.920}$ & $-$ & $-$ & $-$
		\\
		SAL & $-$ & $-$ & $0.785$ & $0.788$
		\\
		FGGAN & $-$ & $-$ & $0.773$ & $-$
		\\ \midrule
		MaliGAN & $-$ & $0.741$ & $-$ & $-$
		\\
		JSD-GAN & $-$ & $0.536$ & $\mathbf{0.894}$ & $\mathbf{0.943}$
		\\ \bottomrule
		\caption{BLEU-2 benchmark regarding architectures that use such a metric to evaluate the four most common datasets.}
		\label{t.benchmark_bleu}
	\end{tabularx}
\end{table}

\subsection{Text Generation using GANs and Gumbel-Softmax Differentiation}
\label{ss.text_gumbel}

Continuous distributions that resemble categorical distributions, such as the Gumbel-Softmax, tackled the differentiation problem and became alternatives to enable proper backpropagations of GANs' gradients. Initially, Kusner and Hernández-Lobato~\cite{Kusner:16} presented the Gumbel-Softmax Generative Adversarial Network (GSGAN), a text-based GAN that uses the Gumbel-Softmax distribution for generating its outputs. Their work was conducted using $5,000$ samples with a maximum length of $12$ characters generated from a context-free grammar (CFG) and evaluated by visual inspection of the generated data compared with a standard Maximum Likelihood Estimation (MLE) pre-trained LSTM. Nevertheless, such an architecture could only produce comparable results to the pre-trained LSTM and was no match for the Relational Generative Adversarial Network (RelGAN), introduced by Nie et al.~\cite{Nie:19}.

The RelGAN introduced a Relational Memory-based generator, a Gumbel-Softmax relaxation, and a multi-embedded representation discriminator to compose their architecture. Their experiments were conducted using synthetic data generated by an oracle LSTM~\cite{Yu:17} and real datasets, such as COCO Image Captions and EMNLP2017 WMT News, being compared to state-of-the-art architectures (MLE, SeqGAN, RankGAN, and LeakGAN). Its main advantage lies in using a Relational Memory generator, which better extracts long-term dependencies due to self-attention layers. In contrast, its main disadvantage lies in needing a proper pre-training to generate feasible text, increasing the overall system's computational burden.

Nonetheless, adversarial-based models' increasing problem was the mode collapsing, where generators tend to sacrifice diversity for quality. In such a context, the Meta Cooperative Training Paradigm Generative Adversarial Network (Meta-CoTGAN), proposed by Yin et al.~\cite{Yin:20}, attempted to slow the effect of mode collapsing. The authors evaluated the impact of the cooperative training language model and meta optimization using the same protocol proposed by Nie et al.~\cite{Nie:19}. Although the authors eased the mode collapsing problem, Meta-CoTGAN had a higher computational cost due to being trained with an additional language model during the adversarial training.

With the mentioned architecture in mind, it is possible to observe that their main disadvantage lies in pre-training both discriminator and generator ahead of the adversarial training, burdening the computational powers. On the other hand, one can perceive that Gumbel-Softmax distributions employ an additional temperature parameter, which controls the generated text's diversity and quality, thus being an interesting parameter to be fine-tuned. Additionally, such models mostly need a mathematical replacement in their output layer (from Softmax to Gumbel-Softmax), which provides a straightforward implementation and a not so complex system.

\subsection{Text Generation using GANs and Reinforcement Learning}
\label{ss.text_rl}

Most Gumbel-Softmax-based approaches have a pre-training burden in advance to the adversarial training and directly rely on traditional GANs objectives, which may cause premature collapsing and an inadequate equilibrium between generator and discriminator. An alternative to tackling such a problem is to bypass the generator's differentiation problem using stochastic gradient policy updates. The idea is to acquire rewards signals from the discriminator and pass them back to the generator in the same way a Reinforcement Learning method would do, such as the REINFORCE algorithm.

The first RL-based work has been introduced by Yu et al.~\cite{Yu:17}, denoted as Sequence Generative Adversarial Network (SeqGAN). Their proposed architecture uses the REINFORCE algorithm, and Monte Carlo searches to propagate the discriminator's gradients to the generator. SeqGAN was the first RL-based work and could only be compared to standard MLE-based training, yet it established a hallmark for future RL-based research. Although SeqGAN tackled adversarial-based text generation, the model lacked generalization power and produced text with minimum diversity. With that in mind, an alternative suggested by Wang et al.~\cite{Wang:17} employs additional features learned by Variational AutoEncoders to represent text better. The so-called VGAN proposed to use high-level latent random variables to model the variability of text, aiding traditional Recurrent Neural Networks in learning well-structured data. Their combined model used an RL-based approach to update the generator's gradients and the same gradient policy defined by SeqGAN. Finally, VGAN outperformed the compared benchmarks, such as recurrent-based Language Models and SeqGAN, over three public literature datasets.

Another SeqGAN extension, denoted as Objective-Reinforced Generative Adversarial Networks (ORGAN), attempted to improve the learning process by including domain-specific objectives in addition to discriminator's rewards. In other words, ORGAN~\cite{Guimaraes:18} proposed to extend the reward function through a linear combination between the discriminator and domain-specific objectives rewards, followed by a Wasserstein distance training. Their experimental results outperformed standard baseline approaches, such as MLE and SeqGAN, while the authors concluded that ORGAN could generate domain-based data and that RL plays an essential role in the model's learning.

Notwithstanding, such precedents enabled researchers to pursue new forms of inputting external information and improving domain-related text generation. In such a context, Li et al.~\cite{Li:18} proposed the Category Sentence Generative Adversarial Network (CS-GAN), which extends traditional text-based GANs with additional category information. Essentially, they proposed both generator and discriminator that incorporate the sentence category (label), allowing the system to capture such information in its learning procedure and generate text according to their category. On the other hand, Rizzo and Van~\cite{Rizzo:20} proposed an extension of the standard SeqGAN by employing context adaptation of global word embeddings and a self-attention discriminator. The so-called C-SeqGAN aimed at using the contribution of global knowledge adapted to the dataset's domain and a self-attentive discriminator that slowly minimizes the loss function and better provides feedback to the generator's updates. The experiments were performed under an extensive evaluation regarding short-, medium, and long-length text datasets and outperformed different SeqGAN-based baselines.

One can perceive the advantage of employing additional information, which assists the network in learning domain-related structures and producing more feasible texts. Nevertheless, using domain-related information may result in substantial mode collapses as the network will lose its diversity capability and generate similar text. Some authors opted not to rely on external information and employ variable reward signals to overcome such a problem. For instance, the Generative Adversarial Network (RankGAN)~\cite{Lin:17} employs a ranking-based architecture to produce high-quality generations. It evaluates and ranks collections of human- and machine-written sequences over a reference group instead of training a discriminator to learn binary predictions. One can perceive that such an approach is inspired by the ranking steps commonly used in web searches and allows the model to calculate the expectations given different references sampled across the reference space. To calculate such values, the authors formulated a relevance score using cosine similarity followed by a softmax activation, which allowed them to compute a collective ranking score for an input sequence. Alternatively, the Diversity-Promoting Generative Adversarial Network (DP-GAN)~\cite{Xu:18} has been proposed to overcome the problem of generating repetitive and ``boring" text sequences. Essentially, DP-GAN assigns a low reward value when generating text repeatedly while assigning high rewards when generating novel text. One can observe that such an approach encourages the generator to produce a more diverse text. Additionally, the authors proposed a new discriminator, which is better at distinguishing between repeated and novel text. 

\begin{sloppypar}
Moreover, the Boundary-Seeking Generative Adversarial Network (BGAN)~\cite{Hjelm:18} uses estimated difference measures from the discriminator to compute importance weights for the generated samples, providing a policy gradient for training the generator. In other words, BGAN models the strong connection between the discriminator's decision boundaries, allowing it to work with continuous data. BGAN uses $f$-divergences, such as Jensen-Shannon, Kullback-Leibler (KL), Reverse KL, or Squared-Hellinger policy gradients and rewards for the generator, being trained with the REINFORCE algorithm. BGAN yielded stable training even though its generation was inadequate as no pre-training was involved. The authors stated that they were not aware of any previous work that could successfully train from scratch a non-continuous architecture and that, although their generation was inadequate, the results demonstrated the stability and capacity of BGANs working with discrete data.
\end{sloppypar}

A recurrent disadvantage of modifying the rewards signals occurs when networks can not provide feasible signals during the starting epochs, hampering the learning process and preventing the generator from being adequately trained. In such a context, few works attempted to present novel RL-based strategies based on pre-training, joint rewards, and additional policy gradient algorithms. Liu et al.~\cite{Liu:19} presented a novel architecture for generating lexically-constrained sentences, denoted as Backward and Forward Generative Adversarial Network (BFGAN). Specifically, their idea was to employ both backward and forward generators with a discriminator that guides their joint training by assigning reward signals, being pre-trained with MLE before the adversarial learning, and then trained using the REINFORCE algorithm.

Regarding text-based GANs, policy gradient methods usually provide reward functions that sample a token per timestep without considering their surroundings and being unfeasible when dealing with long sequences. Thus, Balagopalan et al.~\cite{Balagopalan:18} introduced a comparative study between three policy gradient alternatives, i.e., REBAR, RELAX, and slightly different REINFORCE. The proposed methods denoted as REGANs, aim at providing distinct reward functions to attain a probability distribution over all tokens generated by the generator. Experiments were conducted using synthetic context-free grammar and compared across REBAR, RELAX, and REINFORCE architectures. Alternatively, an Inverse Reinforcement Learning~\cite{Ziebart:08} algorithm has been proposed for text-based GANs, denoted as IRL~\cite{Shi:18}. The authors stated that their method has two main advantages over the traditional RL algorithm: (i) the reward function can produce denser reward signals, and (ii) an entropy regularized-based policy gradient encourages the generation of diverse texts. Unlike traditional GANs, IRL does not directly use a discriminator, where, instead, it uses a reward function that aims to increase the rewards from authentic texts and decrease the rewards from the generated texts. Thus, the generator is trained via entropy regularized policy gradients~\cite{Nachum:17} by maximizing the expected rewards. The experimental results attest that the proposed method alleviated reward sparsity and mode collapsing while achieving comparable results regarding the previous state-of-the-art, e.g., LeakGAN.

Most approaches mentioned above struggled to generate human-like texts and often faced a lack of diversity in the created sentences. With that in mind, some authors attempted to mimic human-like text generation by using previously unused features, such as variable text length, emotion label (positive, negative, or neutral), and controllable topic. The Conditional Text Generative Adversarial Network (CTGAN)~\cite{Chen:19} is trained using the REINFORCE algorithm and composed of a conditional LSTM generator that uses the emotion label and the text as its input. Additionally, it employed a conditional discriminator (standard CNN) to classify whether the text is real or generated. The authors customized the text generation using specific keywords and automated the word-level replacement strategy to extract the target word and replace it with a similar one.

Instead of only using emotions as additional features, Wang and Wan~\cite{Wang:19} extended the previous work by proposing both Sentimental Generative Adversarial Network (SentiGAN) and Conditional Sentimental Generative Adversarial Network (C-SentiGAN), which are capable of generating text with different sentiment labels. In short, the idea is to train multiple generators simultaneously and use a penalty-based objective function to force each generator in sampling specific sentiment texts. Furthermore, a multi-class discriminator helps each generator in generating its particular texts. Another architecture capable of generating emotional text is the Cross-Domain Text Sentiment Transfer Generative Adversarial Network (CD-GAN)~\cite{Zhang:19}, which uses annotated data from other domains and overcomes the problem of large-scale annotated data requirement. The CD-GAN aimed to combine adversarial RL and supervised learning to extract sentiment transformation patterns and generate emotional text. Its architecture comprises a Seq2Seq generator and two discriminators, one accountable for sentiments and another for domains. Both discriminators outputted rewards to update the generator's gradients and enabled the generator to observe how good its generation is based on the input data sentiment (non-labeled data) and related to the auxiliary domain (labeled data).

An alternative to traditional Reinforcement Learning methodologies is to split their knowledge into additional modules, such as manager/worker and actor/critic. Such architectures' advantage is their capacity to represent the policy functions better as they consider past actions and compare them to baselines to provide the reward signals. In other words, it decides how much reward should be given to a generator based on its previous text generations, guiding the network to more feasible updates and hence, better quality text. With that in mind, Guo et al.~\cite{Guo:18} proposed the Leaked Information Generative Adversarial Network (LeakGAN), which allows the discriminator to leak high-level extracted features to the generator and helps to guide the generator as it provides additional information through its learning steps. The idea is to split the generative network into two modules, the MANAGER and WORKER, where the MANAGER takes the leaked information from the discriminator and builds a feature vector further to feed the WORKER module. Both MANAGER and WORKER modules are distinct RNN-based architectures and trained via the REINFORCE algorithm. Additionally, the authors proposed a Bootstrapped Re-scaled Activation, where rewards are re-scaled to mitigate the vanishing gradient problem. The experimental results showed that leaking information is vital to the network's learning process and better guides the generator in generating long text, achieving performance improvements over the previous state-of-the-art architectures.

Following the actor-critic-based architectures, Fedus et al.~\cite{Fedus:18} introduced the Mask Generative Adversarial Network (MaskGAN), which uses an actor-critic conditional GAN to fill in missing text conditioned on the surrounding context. One can perceive that such a task is more challenging than text generation, helping the model reduce the generator's risk competing with an almost-perfect discriminator. Additionally, in-filling tasks might mitigate the problem of mode collapsing as they provide rewards every timestep. The generator is based on a Seq2Seq~\cite{Sutskever:14} architecture and is trained via policy gradients using the REINFORCE algorithm. Furthermore, rewards are generated at each timestep, and the maximum sequence length is increased upon satisfying a convergence criterion. Also, to alleviate the REINFORCE variance problem, the authors proposed generating rewards based on the whole generator's distribution instead of only using the sampled token. On the other hand, Zhang et al.~\cite{CZhang:19} improved the standard actor-critic methodology by proposing a transformer-based architecture for the generator, denoted as Transformer Generative Adversarial Network (TranGAN). Essentially, an encoder-decoder system constitutes the transformer-based architecture, where each layer is composed of multi-head self-attention and feed-forward networks in an attempt to extract long-term dependencies and information from the surrounding context. The proposed model, TranGAN, was compared against a standard Seq2Seq architecture and SeqGAN, outperforming both models in almost all metrics, while the authors concluded that they could effectively process discrete sequences through the actor-critic training algorithm.

Despite already known several RL-based strategies, researchers often grappled with the incapacity of generating long texts within a feasible amount of computation burden. Additionally, the lack of context (text semantics over time) in most GAN-based text generation architectures fostered the necessity of finding alternative improvements. Therefore, Sun et al.~\cite{Sun:19} presented a Quasi Generative Adversarial Network (QuGAN) for Tibetan Question Answering corpus generation, which extends traditional GANs by using the power of Quasi-Recurrent Neural Networks (QRNN) and improving both reward and Monte Carlo search strategies. As QRNNs are combinations of both RNNs and CNNs, they are often used to deal with long sequences of data and data parallelization, providing a faster text generation. Additionally, the authors employed a BERT model after the text generation to correct the generator's grammatical mistakes. Alternatively, by employing a truth-guided generator and self-attention-based discriminators, Wu and Wang~\cite{Wu:20} proposed the Truth-Guided Sequential Generative Adversarial Network (TG-SeqGAN) to improve text generation and make it look closer to the real data. The generator is essentially rewarded with the distance between real and fake data, guiding its convergence, while the discriminator network has a self-attention mechanism to capture long-term dependencies.

Another general improvement, proposed by the Self-Adversarial Learning (SAL)~\cite{Zhou:20} architecture, is to employ a comparative discriminator capable of comparing text quality among samples' pairs. Moreover, throughout the training step, SAL rewards the generator when a generated sentence appears to be better than the previously generated ones. Such an approach stands for a self-improvement reward mechanism, which allows the model to receive more valuable feedback and avoid collapsing due to the limited number of real samples, thus reducing reward sparsity and the risk of mode collapse.

Finally, some new works attempt to improve the text generation quality by applying guidance functions to the discriminator's signals. Essentially, the idea is to provide more complex feedback and, consequently, a more feasible reward signal from the discriminator to the generator, leading to better and more diverse text. Yang et al.~\cite{Yang:20} introduced the Feature-Guiding Generative Adversarial Network (FGGAN), which solves the insufficient feedback guidance from the discriminator through a feature guidance module. The module is a CNN that considers both generated and real data and performs a series of convolutional operations to extract a high-order feature vector and provide such a vector in the form of a reward signal. Additionally, the generated tokens pass through semantic rules to remove illogical tokens and improve the generated text's quality.

At the same time, the TextGAIL~\cite{QWu:20} attempts to improve the discriminator's guidance by combining large-scale pre-trained language models, i.e., RoBERTa and GPT-2, with recent advancements in Reinforcement Learning, such as Generative Information Learning (GAIL) and proximal policy optimization (PPO). The TextGAIL architecture extends the traditional adversarial learning with GAIL and introduces an imitation replay method to stabilize the training, i.e., vocabulary sizes are often large and hampers the GAIL stability. Additionally, ground truth sequences are filled in when training the generator, occasionally forcing constant rewards and stabilizing the generator's convergence. Their discriminator uses a pre-trained RoBERTa model as the final classifier and uses a contrastive-based approach that estimates the relative reality between generated and real sequences, i.e., the discriminator estimates how much a real sentence is more realistic than a generated sentence. As the discriminator is not stationary during adversarial learning, the generator suffers from high variance gradients and needs a type of policy optimization to stabilize its convergence, such as the PPO.

\subsection{Text Generation using GANs and Modified Training Objectives}
\label{ss.text_modified_training}

An alternative to continuous distributions and Reinforcement Learning is to propose new training objectives, hence withdrawing the necessity of differentiable Softmax-like, policy reward functions, and pre-training pipelines. The first work that attempted to introduce such a novel was Che et al.~\cite{Che:17}, which presented the Maximum-Likelihood Augmented Generative Adversarial Network (MaliGAN). Essentially, they introduced a novel objective based on the normalized maximum likelihood and importance sampling, which resembles an MLE-like objective and brings more stability to the training procedure. Additionally, one can observe that their variance reduction techniques assisted create discriminators capable of detecting whether a generator has learned too much noise.

A different architecture that proposes variance reduction techniques and aims at stabilizing the training procedure is the Improved Wasserstein Generative Adversarial Network (IWGAN)~\cite{Gulrajani:17}, which uses a gradient penalty technique instead of the standard weight clipping. Wasserstein networks commonly use weight clipping to stabilize their training, yet in a hard-optimization task, such a technique may not converge under a specific clipped value. In contrast, a gradient penalization is an alternative method to enforce the Lipschitz constraint by introducing a penalization factor to the critic's (discriminator) loss function. Unfortunately, there were not enough baselines to compare and attest whether IWGAN would scale with larger language models.

The literature also started to employ curriculum learning to mitigate the variance noise, where the algorithm slowly teaches the model how to generate increasing and variable-length sequences. Such a model denoted as AltGAN~\cite{Press:17} uses a continuous relaxation mechanism, where the softmax outputs are multiplied with their respective embeddings to provide a differentiable generator. Their experimental results attested that their model could learn from scratch without resorting to any pre-training, being the first model to accomplish such a fact.

Another alternative approach to modify the training objective is by encoding sentences in latent feature distributions similar to AutoEncoders. The first work in such a category has been proposed by Zhang et al.~\cite{Zhang:17}, which introduced a text-based Generative Adversarial Network (TextGAN) that matches the high-dimensional latent feature distribution of real and fake sentences using a kernelized discrepancy metric known as Reproducing Kernel Hilbert Space (RKHS). Such a model used a soft-argmax operator~\cite{Zhang:16} as well as several complementary techniques, such as initialization strategies and discretization approximations, to alleviate the mode collapse problem. Donahue and Rumshisky~\cite{Donahue:18} extended the previous work and proposed a latent-space text-based Generative Adversarial Network (LaTextGAN), which uses an AutoEncoder and low-dimensional latent spaces. In short, they attempt to remove the dimensionality curse and provide an encoding-decoding system that can generate low-dimensional latent-based spaces to discriminate the sequences better. Some plotted t-SNE embeddings showed that LaTextGAN could adequately model the latent space, while interpolated sentences indicated that LaTextGAN could smoothly encode sentences in the latent space. 

Ahamad~\cite{Ahamad:19} introduced a Skip-Thought Vector Generative Adversarial Network (STGAN), which uses an encoder-decoder architecture to generate sentence embeddings. The proposed method does not resort to soft-differentiation or RL, as it generates text by feeding continuous embedded samples. Additionally, it did not have a specific training objective, which allowed the authors to compare and benchmark several training algorithms, such as mini-batch discrimination, gradient penalty, WGAN, and WGAN-GP. Following the STGAN proposal, the Text Knowledge Distillation Generative Adversarial Network (TextKD-GAN)~\cite{Haidar:19} also aimed at producing continuous text representation to replace the traditional one-hot encodings. Instead of only producing embeddings with AutoEncoders, the TextKD-GAN resorted to a teacher-student model, feeding the data through an AutoEncoder and further reconstructing it. Such an approach provides a more challenging problem to the discriminator and increases the generator's ability to fool the discriminator.

Alternatively, the literature suggested using distance-based functions instead of encoder-decoder architectures to provide differentiable and straightforward objectives. In such a context, the Feature Mover Generative Adversarial Network (FM-GAN)~\cite{Chen:18} matches the latent feature distributions of real and fake sentences using Feature-Mover's Distance (FMD), leading to highly discriminative critics and RL-free approaches. The FMD is based on a variation of the Earth-Mover's Distance (EMD), which can be solved by the Inexact Proximal algorithm and further be applied to model discrete data objectives. This architecture could produce good sentences but still suffered from mode collapse (low diversity), while the compared baselines produced high diversity yet low-quality sentences. In such a context, Li et al.~\cite{Li:19} presented an alternative distance-based architecture, the Jensen-Shannon Divergence Generative Adversarial Network (JSD-GAN). Such a model optimizes the Jensen-Shannon Divergence between the generator's distribution and the empirical distribution over the training data, i.e., an alternative to the min-max optimization, which uses a closed-form solution for the discriminator. Although their architecture provides an RL-free algorithm, they could only apply it to explicit representations, thus hindering the number of benefited tasks.

In contrast, some authors found out that using closed-form solutions directly optimizes the training objectives. For instance, Montahaei et al.~\cite{Montahaei:19} introduced the Discrete Generative Self-Adversarial Network (DGSAN), which uses an optimized closed-form solution to find an optimal discriminator. Essentially, the DGSAN combines both generator and discriminator powers in a single network, where the information is encoded into a new objective that creates a relationship between the current generator, the new generator, and the optimal discriminator between them. In short, the proposed method takes advantage of not having to calculate gradients, allowing for a direct sampling over an explicit distribution.
\section{Conclusions and Future Insights}
\label{s.conclusion}

Throughout the last years, increasing demand for understanding and automating human communication fostered Natural Language Processing research. Machine Learning and abundant computational power helped achieve several hallmarks in a wide range of NLP-based tasks, thus becoming viable alternatives for modeling human-machine interactions.

Nevertheless, such systems still struggle when dealing with adversarial or unknown data, i.e., data that has been slightly modified or not present in the original one. On the other hand, one can perceive that such a problem encourages researchers to find feasible alternatives, such as Adversarial Learning. In essence, Adversarial Learning attempts to overcome the problem of adversarial data by introducing noisy examples to an architecture's learning process, aiding in its robustness and performance. Moreover, since the advent of Generative Adversarial Networks, it is now possible to learn the data's distribution and generate artificial data that resembles the original one.

This article presented a survey on the most recent studies concerning text generation using Generative Adversarial Networks. This paper's most significant contribution is to critically analyze and provide a unique source of recent GAN-based text generation research, mostly ranging from 2016 to 2020. Additionally, Table~\ref{t.summary} presents an overview of the considered works, easing the reader's needs when looking for particular researches.

We have observed that GAN-based text generation works are brand new (between 2016 and 2020); hence they still struggle with a lack of researches and adequately defined pipelines. GAN-based text generation's primary challenges are that GANs were not designed to work with discrete data, needing additional tricks to subdue such a problem. Such data do not provide differentiable outputs, which inhibits gradients from being appropriately calculated and updated. Therefore, most works attempt to override such a problem by employing modified training objectives, Reinforcement Learning, or continuous-based outputs, such as Soft-Argmax or Gumbel-Softmax distributions.

Another barrier regards the intrinsic features of a language, i.e., grammar, syntax, and semantic properties. To provide a feasible text generation, the model needs to learn how characters and words connect between themselves, often accomplished by sorts of memories and contexts (prior knowledge). We believe that such problems can be dealt with in a more robust pre-learning step, in which pre-trained embedding models guide networks instead of MLE, such as BERT~\cite{Devlin:18}, ELECTRA~\cite{Clark:20}, and GPT-2~\cite{Radford:19}, among others. 

The advent of Transformer-based architectures, which provides substantial methods that can generate plausible ``natural"~language, initially diverted GAN-based systems' attention; however, their outstanding performance might benefit GANs when applied as pre-training approaches or even as part of their architectures. With that in mind, it seems that there are loads of blank spots that still need to be appropriately addressed by the research community regarding GAN-based text generation, whereas Transformer-based GANs look like promising approaches to be conducted in 2021.

%

\section*{Acknowledgments}

\begin{sloppypar}
The authors are grateful to S\~ao Paulo Research Foundation (FAPESP) grants \#2013/07375-0, \#2014/12236-1, \#2019/07665-4, \#2019/02205-5, and \#2020/12101-0, and to the Brazilian National Council for Research and Development (CNPq) \#307066/2017-7 and \#427968/2018-6.
\end{sloppypar}

\bibliography{references}

\end{document}